\documentclass[fleqn]{article}

\usepackage[preprint]{nips_2018}

\usepackage[utf8]{inputenc} 
\usepackage[T1]{fontenc}    
\usepackage{hyperref}       
\usepackage{url}            
\usepackage{booktabs}       
\usepackage{amsfonts}       
\usepackage{nicefrac}       
\usepackage{microtype}      

\usepackage{amsmath}
\usepackage{amssymb}
\usepackage{color}
\usepackage{natbib}
\usepackage{makecell}
\usepackage{cuted}
\usepackage{comment}
\usepackage[capbesideposition=outside,capbesidesep=quad]{floatrow}

\usepackage{tikz}
\usetikzlibrary{arrows,shapes,backgrounds,through,shadows}

\tikzstyle{cont}=[circle, draw,thick,minimum size=7.5mm,line width=1pt,>=stealth]  

\tikzstyle{obs}=[fill=blue!10,thick]  

\tikzstyle{contobs}+=[cont]
\tikzstyle{contobs}+=[obs]
\tikzstyle{discobs}+=[disc]
\tikzstyle{discobs}+=[obs]

\tikzstyle{dgraph}=[->, line width=1.5pt]

\usepackage{amssymb,array,delarray,verbatim,alltt,epsfig,float,amsthm}
\usepackage{graphicx,xspace}
\usepackage{url}
\usepackage{amssymb,amsmath}
\usepackage{pgf,amsbsy,amsfonts,amsmath,subfigure,multicol,natbib}
\usepackage{graphicx,xspace}
\usepackage{tikz}
\usetikzlibrary{arrows,shapes,backgrounds,through,shadows}
\usetikzlibrary{decorations.pathmorphing}
\usetikzlibrary{calc}
\usetikzlibrary{matrix}

\newcommand{\mydelta}{\epsilon}

\tikzstyle{bigdot}=[circle,draw=blue!80,fill=blue!25,anchor=center,align=center,minimum size=6mm]
\tikzstyle{bigemptydot}=[circle,draw=blue!80,anchor=center,align=center,minimum size=6mm]

\tikzstyle{biggreendot}=[circle,draw=green!80,fill=green!25,anchor=center,align=center,minimum size=6mm]
\tikzstyle{bigreddot}=[circle,draw=red!80,fill=red!25,anchor=center,align=center,minimum size=6mm]

\tikzstyle{dot}=[circle,draw=orange!80,fill=orange!25,anchor=center,align=center,minimum size=4mm]
\tikzstyle{emptydot}=[circle,draw=orange!80,anchor=center,align=center,minimum size=4mm]

\tikzstyle{reddot}=[circle,draw=red!80,fill=red!25,anchor=center,align=center,minimum size=4mm]

\tikzstyle{greendot}=[circle,draw=green!80,fill=green!25,anchor=center,align=center,minimum size=4mm]

\tikzstyle{bluedot}=[circle,draw=blue!80,fill=blue!25,anchor=center,align=center,minimum size=4mm]

\tikzstyle{neur}=[rectangle,draw=green!50,fill=green!50,minimum size=6mm,line width=2pt,>=stealth]  

\tikzstyle{fact}=[fill,minimum size=1.5mm,line width=2pt,>=stealth]

\tikzstyle{cont2}=[circle,draw=black!50,top color=white, 
bottom color=lilla!50!black!20,thick,minimum size=6mm,line width=1pt,>=stealth]  

\tikzstyle{contredb}=[circle,draw=red,top color=red, 
bottom color=red!50!black!20,thick,minimum size=8.1mm,line width=1pt,>=stealth]  
\tikzstyle{contyellowb}=[circle,draw=yellow,top color=yellow, 
bottom color=yellow!50!black!20,thick,minimum size=8.1mm,line width=1pt,>=stealth]  
\tikzstyle{contblueb}=[circle,draw=blue,top color=blue, 
bottom color=blue!50!black!20, thick,minimum size=8.1mm,line width=1pt,>=stealth]  

\tikzstyle{contred}=[circle,draw=red,top color=red, 
bottom color=white,thick,minimum size=6mm,line width=1pt,>=stealth]  
\tikzstyle{contyellow}=[circle,draw=yellow,top color=yellow, 
bottom color=yellow!50!black!20,thick,minimum size=6mm,line width=1pt,>=stealth]  
\tikzstyle{contblue}=[circle,draw=blue,top color=blue!80!black!50, 
bottom color=white, thick,minimum size=6mm,line width=1pt,>=stealth]  
\tikzstyle{contgreen}=[circle,draw=green,top color=green!80!black!50, 
bottom color=white, thick,minimum size=6mm,line width=1pt,>=stealth]  
\tikzstyle{contwhite}=[circle,draw=white,color=white, thick,minimum size=6mm,line width=1pt,>=stealth]  
\tikzstyle{contwhiteb}=[circle,draw=white,color=white, thick,minimum size=7.5mm,line width=1pt,>=stealth]  

\tikzstyle{cont}=[circle, draw,
thick,minimum size=6mm,line width=1pt,>=stealth]  

\tikzstyle{contb}=[circle,draw=black!50,top color=white, 
bottom color=darkgreen!50!black!20,thick,inner sep=0pt,minimum size=8.1mm,line width=1pt,>=stealth]  

\tikzstyle{ocont}=[ellipse,draw=blue!50,thick,minimum size=6mm,>=stealth]  
\tikzstyle{blackcont}=[circle,draw=black!50,thick,minimum size=6mm,line width=2pt,>=stealth]  
\tikzstyle{oval}=[ellipse,draw=blue!50,thick,minimum size=6mm,line width=1pt,>=stealth]  
\tikzstyle{ovalb}=[ellipse,draw=blue!50,thick,minimum size=7.5mm,line width=1pt,>=stealth]  
\tikzstyle{disc}=[rectangle,draw=blue!50,thick,line width=1pt,minimum size=6mm]  
\tikzstyle{obs}=[fill=blue!20,thick]  
\tikzstyle{opt}=[star,draw=red!50,thick,minimum size=6mm]  

\tikzstyle{fillred}=[fill=red!20,thick]  
\tikzstyle{fillgreen}=[fill=green!20,thick]  
\tikzstyle{purered}=[fill=red]  
\tikzstyle{state}=[rectangle,fill=red!20]  
\tikzstyle{sobs}=[fill=green!15,thick]  
\tikzstyle{fact}=[fill,minimum size=1.5mm,line width=2pt,>=stealth]
\tikzstyle{varfact}=[draw,minimum size=1.5mm,line width=2pt,>=stealth]
\tikzstyle{sep}=[rectangle,draw=magenta!50,thick,minimum size=6mm]  
\tikzstyle{det}=[fill=red!15,rectangle,draw=red!50,thick,minimum size=6mm]  

\tikzstyle{dethid}=[diamond,draw=red!50,thick,minimum size=6mm]  

\tikzstyle{lineball}=[fill,-*,draw=red!50,line width=1.5pt]
\tikzstyle{redball}=[mark=*,mark options={fill=red!50,draw=red},mark size=0.5pt]
\tikzstyle{greenball}=[mark=*,mark options={fill=green!50,draw=green},mark size=0.5pt]
\tikzstyle{hid}=[circle,draw,thick]  

\tikzstyle{dec}=[rectangle,draw=red!50,thick,minimum size=6mm]  
\tikzstyle{utility}=[diamond,draw=red!50,thick,minimum size=6mm]  
\tikzstyle{contdec}=[circle,draw=blue!50,thick,fill=blue!10,line width=2pt]  
\tikzstyle{decutility}=[diamond,draw=red!50,thick,minimum size=6mm]  

\tikzstyle{contobs}+=[cont]
\tikzstyle{contobs}+=[obs]
\tikzstyle{discobs}+=[disc]
\tikzstyle{discobs}+=[obs]

\tikzstyle{obsred}+=[obs]
\tikzstyle{obsred}+=[red]

\tikzstyle{background grid}=[draw, black!50,step=.1cm]
\tikzstyle{dgraph}=[->, line width=1.5pt]
\tikzstyle{ugraph}=[line width=1.5pt]

\definecolor{magenta}{cmyk}{0.1,1,1,0.5}
\definecolor{darkgreen}{cmyk}{0.6,0.1,0.6,0.6}
\definecolor{pink}{cmyk}{0.1,1,1,0.1}
\definecolor{azzurro}{cmyk}{0.9333, 0.2471, 0.5569, 0.102}
\definecolor{lilla}{rgb}{0.5, 0, 0.5}
\definecolor{mygreen}{rgb}{0, 0.5, 0}
\definecolor{darkorange}{rgb}{1, 0.4, 0} 
\definecolor{darkred}{rgb}{0.8, 0, 0}

\newcommand{\eframe}{\end{frame}}
\newcommand{\bframe}[1]{\begin{frame}{#1}}

\newcommand{\tw}{\textwidth}

\newcommand{\bmp}[1]{\begin{minipage}{#1}}
\newcommand{\bmpp}[2]{\begin{minipage}[#1]{#2}}
\newcommand{\emp}{\end{minipage}}

\newcommand{\br}[1]{\left( {#1} \right)}
\newcommand{\sq}[1]{\left[ {#1} \right]}
\newcommand{\rbar}[1]{\left. {#1} \right\rvert}

\newcommand{\m}[1]{\mathbf{#1}}

\newcommand{\ave}[1]{\mathbb{E}{\sq{#1}}}

\newcommand{\Id}{\m{I}}

\newcommand{\sett}[1]{\myset{#1}}

\newcommand{\hcm}{\hspace{0.25cm}}

\newcommand{\myset}[1]{\mathcal{\uppercase{#1}}}

\newcommand{\beq}{\begin{equation}}
\newcommand{\eeq}{\end{equation}}

\tikzstyle{celim}=[circle,draw=red!25,thick,minimum size=6mm,line width=2pt,>=stealth]  %
\tikzstyle{delim}=[draw=red!25]  %

\newcommand{\btz}{\begin{tikzpicture}}
\newcommand{\etz}{\end{tikzpicture}}

\newcommand{\ep}{\epsilon}
\newcommand{\epp}{\tilde{\epsilon}}

\newcommand{\trace}[1]{{\rm trace}\left({#1}\right)}

\newcommand{\pdiff}[2]{\frac{\partial{#1}}{\partial{#2}}}

\newcommand{\half}{\frac{1}{2}}

\renewcommand{\eqref}[1]{(\ref{#1})}
\newcommand{\secref}[1]{Section \ref{#1}}
\newcommand{\figref}[1]{Figure \ref{#1}}
\newcommand{\figrefm}[2]{Figure \ref{#1}(#2)}

\newcommand{\Figref}[1]{Figure \ref{#1}}

\def\be{\begin{equation}}
\def\ee{\end{equation}}

\title{Stochastic Variational Optimization}

\author{
  Thomas Bird \\
  University College London\\
  \texttt{thomas.bird.17@ucl.ac.uk} \\
  \And
  Julius Kunze \\
  University College London \\
  \texttt{juliuskunze@gmail.com} \\
  \And
  David Barber \\
  University College London \\
  \texttt{david.barber@ucl.ac.uk} \\
}

\begin{document}

\maketitle

\begin{abstract}
  Variational Optimization forms a differentiable upper bound on an objective. We show that approaches such as Natural Evolution Strategies and Gaussian Perturbation, are special cases of Variational Optimization in which the expectations are approximated by Gaussian sampling. These approaches are of particular interest because they are parallelizable. We calculate the approximate bias and variance of the corresponding gradient estimators and demonstrate that using antithetic sampling or a baseline is crucial to mitigate their problems. We contrast these methods with an alternative parallelizable method, namely Directional Derivatives. We conclude that, for differentiable objectives, using Directional Derivatives is preferable to using Variational Optimization to perform parallel Stochastic Gradient Descent.
\end{abstract}

\section{Introduction}

We consider approaches to minimizing a scalar valued function $f$ with respect to vector argument $x$. The high compute requirements of deep learning models have been the genesis of efforts to parallelize this optimization process \cite{zinkevich2010, dean2012}. Recently there has been interest in parallel, simple gradient approximation methods, in particular what we term the Gaussian Perturbation (GP) estimator \cite{salimans17}.

We make the following contributions:

\begin{itemize}
    \item We describe the Stochastic Variational Optimization approach (SVO), and show that it is a principled form of Evolutionary Optimization based on a simple upper bound. We show how GP \cite{salimans17} is a special case of SVO in which the approximating distribution is a Gaussian.
    \item We calculate the bias and variance of the GP estimator and show that it has high variance, explaining why the training may not converge or become unstable. 
    \item We show antithetic sampling does not reduce bias but does dramatically reduce the variance of the estimator; we derive an approximation for this variance. We demonstrate that using a simple baseline can also achieve this dramatic variance reduction.
    \item We show the close relationship between the GP estimator with antithetic sampling and the SPSA \cite{spall} gradient estimator.
    \item In the case of differentiable objectives we show the existence of an alternative parallel gradient estimator based on Directional Derivatives. We show that this approach has much lower variance than the standard GP approach, with similar (but slightly better) performance to GP with antithetic sampling.
\end{itemize}

\section{Stochastic Variational Optimization}

Variational Optimization is based on the simple observation 
\begin{equation}
\min_x f(x) \leq \ave{f(x)}_{p(x|\theta)}
\end{equation}
where $\theta$ is a set of continuous parameters of the variational distribution $p$. That is, the minimum of a collection of values is always less than their average.  By defining 
\beq
U(\theta) = \ave{f(x)}_{p(x|\theta)}
\eeq
Instead of minimizing $f$ with respect to $x$, we can minimize the upper bound $U$ with respect to $\theta$. In the original VO work \cite{StainesBarber12} the focus was on forming a differentiable upper bound for non-differentiable $f$ or discrete $x$.  

\begin{figure}
\floatbox[{\capbeside}]{figure}[\FBwidth]
{\caption{The original function $f(\mu)$ is plotted in black. The upper bound function $U(\mu)$ is plotted for $\sigma^2=100,20,5$, resulting in the red, magenta and blue curves. As $\sigma^2$ reduces, $U(\mu)$ becomes an increasingly good approximation to $f(\mu)$.}\label{fig:sketch}}
{\scalebox{1.5}{\includegraphics[width=5cm]{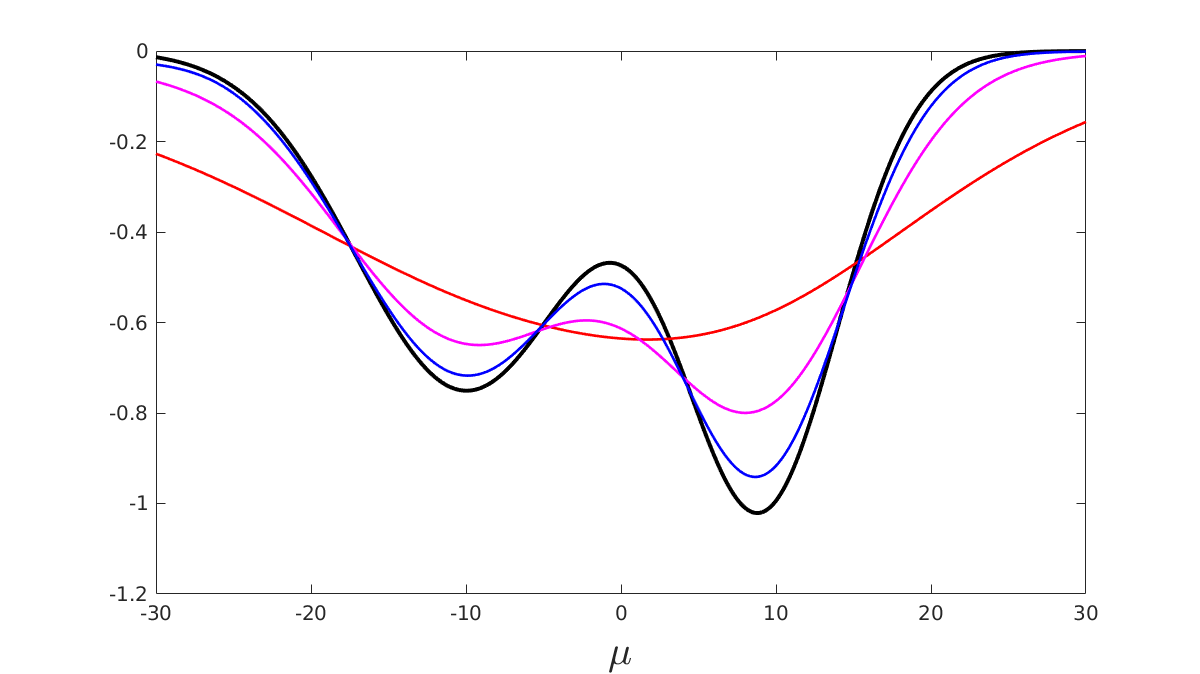}}}
\end{figure}

For example, using a Gaussian distribution with mean $\mu$ and covariance $\sigma^2\Id$ for the variational $p(x|\theta)$, as a function of the mean $\mu$, the upper bound can be written
\beq
U(\mu) = \ave{f(\mu+\sigma z)}_{N(\bf{0},\Id)}
\eeq
Assuming $f$ is smooth and expanding to second order around $\mu$, we obtain
\beq
U(\mu) = f(\mu)+ \frac{\sigma^2}{2}\trace{H} + O(\sigma^4)
\eeq
where $\trace{H}$ is the trace of the Hessian of $f$ evaluated at $\mu$.  This means that, for local minima in $\mu$, $U(\mu)$ lies above $f(\mu)$; hence $U$ is an upper bound on not just the global minimum of $f(\mu)$, but also any local minima. Note that, this does not mean that $U(\mu)\geq f(\mu)$ for all $\mu$. However, any minimum of $f$ is below $U$ and any maximum is above $U$ (since at that point $\trace{H}$ is negative). See Figure \ref{fig:sketch}.

The gradient of the upper bound can be computed by any standard means. However, it is interesting to express it as 
\beq
\frac{\partial U}{\partial \theta} = \ave{f(x)\frac{\partial}{\partial \theta}\log p(x|\theta)}_{p(x|\theta)} \label{reinforce}
\eeq
which is reminiscent of the `reinforce' algorithm \cite{Williams:1992:SSG:139611.139614}.

There is a connection to evolutionary computation (more precisely Estimation of Distribution Algorithms \cite{StainesBarber13, NES}) if the expectation with respect to $p(x\vert \theta)$ is performed using sampling. In this case one can draw samples $x^1,\ldots,x^S$ from $p(x\vert\theta)$ and form an unbiased approximation to the upper bound gradient 
\beq
\frac{\partial U}{\partial \theta} \approx \frac{1}{S} \sum_{s}f(x^s)\frac{\partial}{\partial \theta}\log p(x^s|\theta)
\eeq
We call this approach Stochastic Variational Optimization (SVO).
The `evolutionary' connection is that the samples $x^s$ can be thought of as `swarm members' that are used to estimate the gradient. 


\begin{figure*}[t]
	\begin{center}
		\subfigure[]{\includegraphics[width=0.3\tw]{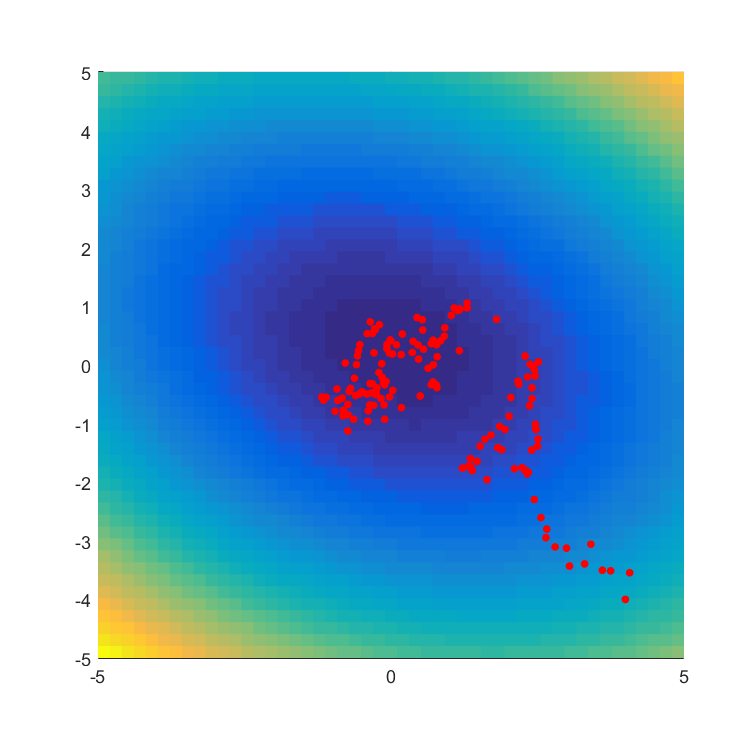}}
		\subfigure[]{\includegraphics[width=0.3\tw]{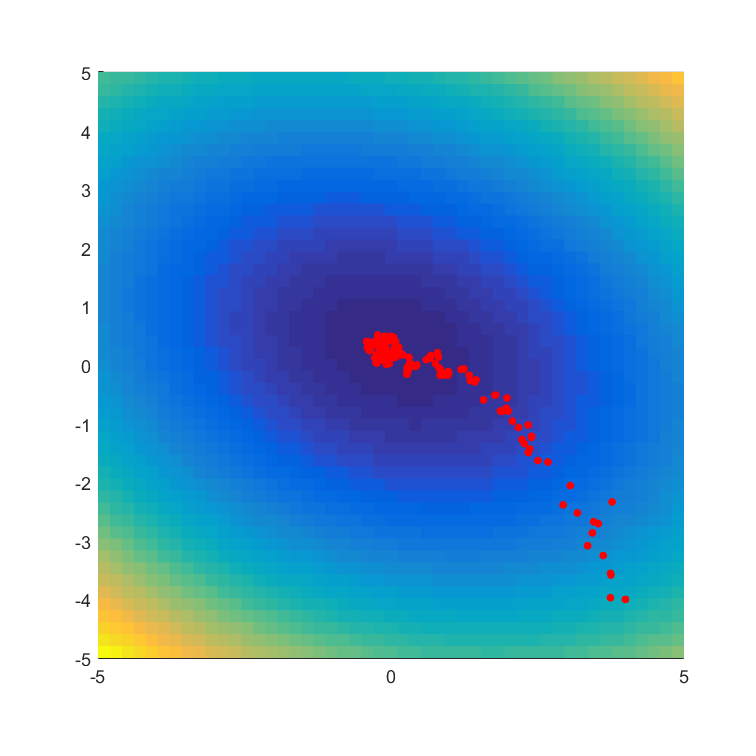}}
		\subfigure[]{\includegraphics[width=0.3\tw]{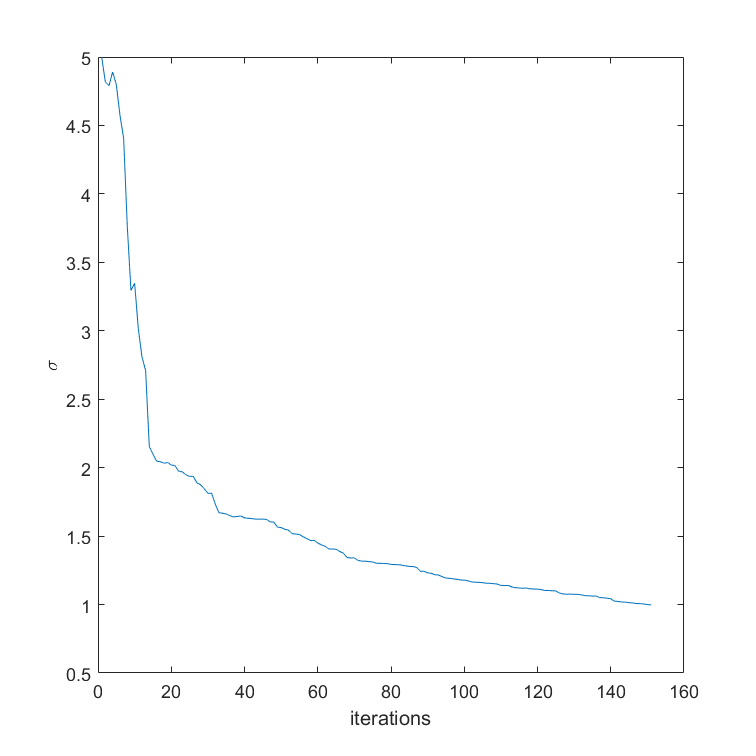}}
	\end{center}
	\caption{Stochastic VO using $S=10$ samples for a quadratic function. (a) We plot the trajectory of the Gaussian mean $\mu$, with the initial parameter in the bottom right. Despite the noisy gradient estimate, the parameter values $\mu$ move toward the minimum of the objective $f(x)$.   (b) Mean trajectory under learning the Gaussian variance $\sigma^2$. (c) Learned $\sigma$ values versus gradient descent iterations. \label{fig:vo}}
\end{figure*}

A special case of VO is to use a Gaussian with the variational parameter $\theta$ being the Gaussian mean $\mu$ so that (the multivariate setting follows similarly)
\beq
U(\mu) = \frac{1}{\sqrt{2\pi\sigma^2}}\int e^{-\frac{1}{2\sigma^2}(x-\mu)^2}f(x)dx
\eeq
where $\sigma^2$ is the (fixed) Gaussian variance. 
%
The gradient of this upper bound is given, after a change of variable $\ep=x-\mu$, by
\beq
U'(\mu) = \frac{1}{\sigma^2}\ave{\ep f(\mu+\ep)}_{\ep \sim N(0,\sigma^2)}
\label{eq:grad}
\eeq
Fixing $\sigma=5$ and using $S=10$ samples, we show in Figure \ref{fig:vo}{a} the trajectory (for 150 steps of SGD with fixed learning rate $\eta=0.1$) of $\mu$ based on Stochastic VO and compare this to the underlying function $f(x)$ (which in this case is a simple quadratic).  

One can also consider $\theta=(\mu,\sigma^2)$ so that the bound is a function of both the mean $\mu$ and variance $\sigma^2$ and minimize the bound with respect to both parameters (parameterizing $\sigma^2=e^\beta$ to ensure a positive variance).  Using a Gaussian with covariance $e^\beta \Id$ and performing gradient descent on both $\beta$ and $\mu$, for the same objective function, learning rate $\eta=0.1$  and initial $\sigma=5$, we obtain the trajectory in Figure \ref{fig:vo}{b}.  As we can see, by learning $\sigma$, the trajectory is much less noisy and more quickly homes in on the optimum.  The trajectory of the learned standard deviation $\sigma$ is given in Figure \ref{fig:vo}{c}, showing how  $\sigma^2$ reduces as we near the minimum\footnote{The reduction of $\sigma$ and homing in on the minimum is somewhat special in this case. The objective chosen here is a quadratic whose minimum value is $0$. As we explain in section \ref{sec:gp:analysis}, for non-zero minimum $f$, convergence cannot be guaranteed.}.

\section{Gradients by Gaussian Perturbation}
In \cite{salimans17} the aim is to minimize a function $f(x)$ without explicitly calculating the gradient. They use the estimator \eqref{eq:grad} for the gradient, corresponding to a choice of a Gaussian for the variational distribution in \eqref{reinforce}. We term this approach the Gaussian Perturbation (GP) approach. 

It's important to stress that in VO the optimization is over the variational parameter $\theta$, not the original variable $x$. The equivalence between VO and GP occurs in the special case of using a Gaussian $p(x|\theta)$  -- the gradient with respect to the mean $\mu$ of the Gaussian is the same as the GP gradient approximation in $x$-space. An advantage of the VO approach, however, is that it provides a principled way to adjust parameters such as the variance $\sigma^2$ (based on minimizing the upper bound).



We will now analyze the properties of this GP gradient estimator, for which we choose our objective function $f$ to be differentiable in order to perform our analysis. It is important to note that whilst $f$ will often be differentiable for a general machine learning setting, the GP estimator is still valid for non-differentiable $f$. In order to perform any meaningful analysis though, some restrictions must be placed on the choice of $f$ to analyze, as the set of all non-differentiable functions is extremely broad. Therefore the relevance of the results we derive for a non-differentiable objective function will be dependendent on how close to differentiable the objective function is. For example a piecewise differentiable $f$ with relatively few pieces will most likely still exhibit the same behaviours we find for smooth $f$.



\subsection{Analyzing the Gaussian Perturbation Estimator\label{sec:gp:analysis}}

For the estimator
\beq
\hat{g}_i =  \frac{1}{S\sigma^2}\sum_{n=1}^S \ep_i^n f(x+\ep^n)
\label{eq:gp:est}
\eeq
where the $\epsilon^n$ are vectors sampled from a zero mean Gaussian with isotropic covariance $\sigma^2\Id$ it is straightforward to calculate the bias (see \secref{sec:svo:details}). The calculation uses the Taylor expansion 
\beq
f(x+\ep)= f(x)+\sum_i \ep_i g_i + \frac{1}{2}\sum_{i,j} \ep_i \ep_j H_{ij} + \frac{1}{3!}\sum_{i,j,k} \ep_i \ep_j \ep_k I_{ijk}+O(\ep^4)
 \label{eq:taylor}
\eeq
where $H$ is the Hessian of second order derivatives and $I$ represents the array of third order derivatives. 
Taking the first two terms of the Taylor expansion gives
\beq
\hat{g}_i =  \frac{1}{S\sigma^2}\sum_{n=1}^S \ep_i^n \br{f(x)+\sum_a \ep^n_a g_a +O(\ep^2)}
\label{eq:gp:bias:expansion}
\eeq
The calculation is straightforward and (expanding to higher order) and taking expectation gives the result
\beq
\ave{\hat{g}_i} =  g_i+\sett{I}_i\sigma^2 +O\br{\sigma^4} 
\eeq
where
\begin{equation}
\sett{I}_i\equiv\frac{1}{2}\br{I_{iii}+\sum_{a\neq i}I_{iaa}}
\label{eq:i_def}
\end{equation}
Except in special cases (for example $f$ has zero third and higher order derivatives), $\hat{g}_i$ is therefore not an unbiased estimator of the gradient $g_i$. As we lower $\sigma^2$ the bias will reduce towards zero.

We can similarly show that the variance $\ave{\hat{g}_i^2}- \ave{\hat{g}_i}^2$ of the estimator is approximately (see \secref{sec:svo:details})
\begin{align}
\frac{1}{S} \bigg( \frac{f^2}{\sigma^2}+\sum_{j} g_j^2+g_i^2 +f\br{\trace{H}+2H_{ii}} + \sigma^2 \bigg( \frac{\sett{H}_i}{4} + \sett{J}_i\bigg) \bigg)+O\br{\sigma^4}
\label{eq:gp:variance}
\end{align}
where  $\sett{H}_i$ is a function of the second derivative of $f$ and $\sett{J}_i$ is a function of the third and first derivatives, see \secref{sec:dd:details}. We note that the terms linear in $f$ can be eliminated from the variance of this estimator by normalizing $f$.

However, the first term, quadratic in $f$, cannot be eliminated by normalization. From this first term in \eqref{eq:gp:variance}, ($f^2/\sigma^2$), we see that there is a tradeoff -- reducing $\sigma$ decreases the bias but increases the variance of the gradient estimator.  Indeed, as we reduce $\sigma$ towards zero, the variance of the gradient estimator can increase without bound. One must therefore use a non-negligible value for $\sigma^2$ in this GP approach, resulting in a bias. In practice the variance of this estimator is therefore very high and one would need many samples to form an accurate estimate of the gradient. 

The average squared error between the true and approximate gradient has, up to $O(\sigma^2)$ the same value as the variance of the gradient estimator. Therefore reducing $\sigma$ can result in a catastrophic increase in the gradient error.

\subsection{Antithetic Sampling}
In antithetic sampling \cite{hammersley, fishman} each sample from a zero mean Gaussian is accompanied by its negative counterpart. So for the GP gradient estimator, the antithetic sampler is defined as
\beq
\sq{\hat{g}_i}_{AS} \equiv \frac{1}{2S\sigma^2}\sum_{n=1}^S \ep_i^n \br{f(x+\ep^n) - f(x-\ep^n)}
\label{eq:as:grad:mean}
\eeq

Using the Taylor expansion of $f$ we get
\beq
\sq{\hat{g}_i}_{AS} \approx \frac{1}{S\sigma^2}\sum_{n=1}^S \ep_i^n\br{\sum_j \ep^n_j g_j +\frac{1}{3!}\sum_{abc} \ep^n_a \ep^n_b \ep^n_c I_{abc}}
\label{eq:as:expansion}
\eeq
The key observation here is that, compared to \eqref{eq:gp:bias:expansion}, the leading term $\ep_i^nf(x)$ is not present. Whilst this term is zero in expectation, this adds considerably to the variance of the standard GP estimator. The bias calculation proceeds as for the standard GP estimator and indeed the bias is exactly the same, namely
\beq
\ave{\hat{g}_i}_{AS} =  g_i+\sett{I}_i\sigma^2 +O\br{\sigma^4} 
\eeq
To calculate the variance of this estimator we can similarly use the existing GP calculation in \secref{sec:svo:details}, and recognize that we can set the terms $f$ and $H$ to zero in \eqref{eq:gp:var:calc} and continue as before. Doing so gives the result that the variance of the GP-AS estimator for $g_i$ is
\beq
\ave{\hat{g}_i^2} - \ave{\hat{g}_i}^2 = \frac{1}{S} \br{\sum_{j} g_j^2 + g_i^2 + \sigma^2\sett{J}_i}+O\br{\sigma^4}
\label{eq:gp:as:variance}
\eeq
Up to $O(\sigma^4)$ this is the same as the expected squared error between the true gradient and the estimated gradient.

Critically, comparing \eqref{eq:gp:as:variance} and \eqref{eq:gp:variance},  the $1/\sigma^2$ term has disappeared in \eqref{eq:gp:as:variance}. Thus, as $\sigma\rightarrow 0$, both the bias \emph{and} variance reduce to zero.
As we converge to the optimum $g_i$ tends to zero, meaning that the variance of the estimator also reduces to zero and becomes independent of $\sigma$.

\subsection{Using a baseline}

We have seen that antithetic sampling is a viable method of reducing the variance in the GP gradient estimator, due to cancellation of normally problematic terms in the Taylor expansion of the estimator. We will now show that a similar cancellation can be achieved by using a simple baseline \cite{greensmith}. Specifically, if we use a baseline of the current function evaluation then we get the following estimator 
\begin{align}
\sq{\hat{g}_i}_{baseline} &\equiv \frac{1}{S\sigma^2}\sum_{n=1}^S \ep_i^n \br{f(x+\ep^n) - f(x)} \label{eq:baseline}\\
&\approx \frac{1}{S\sigma^2}\sum_{n=1}^S \ep_i^n\br{\sum_j \ep^n_j g_j +\frac{1}{2!}\sum_{ab} \ep^n_a \ep^n_b H_{ab}}
\end{align}
Where we have expanded and kept the two leading terms.

We can see that, as with antithetic sampling (\ref{eq:as:expansion}), the $f$ terms cancel. Thus in a similar manner the variance of this estimator will not contain the problematic $f^2/\sigma^2$ term we see in (\ref{eq:gp:variance}), and the variance will be well behaved as $\sigma \rightarrow 0$.

This estimator will have the same bias but a higher variance compared to using antithetic sampling, as the higher order terms (e.g. terms involving $H_{ab}$) will not cancel when using the baseline. However, the baseline does have an advantage that $f(x)$ only has to be calculated once and reused for all samples to evaluate (\ref{eq:baseline}), whereas antithetic sampling requires another $S$ evaluations to calculate the function values at the mirrored samples. Thus we can get a well-behaved gradient estimate using the above baseline using roughly half the number of samples we would use for antithetic sampling.

In practice, we find that the methods perform largely similarly, and we use antithetic sampling for the experiments in section \ref{sec:demos}. 

From this analysis, we see that using antithetic sampling or an appropriate baseline is critical to make the GP estimator practical. Without these, even classical variance reduction methods such as control variates \cite{NES} will not be sufficient to reduce the unbounded variance of the standard GP estimator to a reasonably small value. However, using antithetic sampling or an appropriate baseline, there is no need to use control variates, provided $\sigma$ is set sufficiently small.  Similarly, these techniques can be used within the Stochastic Variational Optimization framework as well, dramatically reducing the variance of the gradient estimator in exactly the same way.

\subsection{Simultaneous perturbation stochastic approximation}

The SPSA, \cite{spall}, gradient estimator is given by
\beq
\hat{g}_i = \frac{1}{2S}\sum_{n=1}^S (\ep_i^n)^{-1}\br{f(x + \ep^n) - f(x - \ep^n)} \label{eq:spsa}
\eeq
In order for the estimator to be valid $\ave{\ep_i^{-1}}$ must be bounded. This rules out using a Gaussian for the perturbation, and the standard choice is to take $\ep_i$ to be Bernoulli distributed and symmetric about 0, e.g. $\ep_i \in \{-\sigma, \sigma \}$. This has expectation and variance
\begin{align}
    \ave{\hat{g}_i} &= g_i + \sigma^2\mathcal{I}_i + O(\sigma^4) \\
    \ave{\hat{g}_i^2} - \ave{\hat{g}_i}^2 &= \frac{1}{S}\br{\sum_{j \neq i}g_j^2 + \sigma^2\mathcal{K}_i} + O(\sigma^4)
\end{align}
where $\mathcal{K}_i$ is a function of the third derivatives. We note that these properties, and also the estimator itself \eqref{eq:spsa}, are closely related to the form of the GP estimator with antithetic sampling. In particular they have the same bias and similar variance. In particular the variance of this estimator, as with GP-AS, reduces as $\sigma \rightarrow 0$. Thus it is a viable method to perform (parallel) approximate gradient calculations.

\section{Stochastic Directional Derivative\label{sec:dd}}

The Directional Derivative (DD) along vector $u$ is the scalar value defined as
\beq
D_u f(x) \equiv \rbar{\pdiff{f(x+\mydelta u)}{\mydelta}}_{\ep=0} = \sum_j u_jg_j
\eeq
The DD can be computed numerically (exactly) by Forward Mode Automatic Differentiation at a cost of approximately two evaluations of the function $f$ \cite{baydin15}. The full gradient can thus be computed by calculating the DD along a set of directions that span the space.  
An estimator for the gradient can be found from a smaller number of directions
\beq
\hat{g} \equiv \frac{1}{S\sigma^2} \sum_{n=1}^S D_{\ep^n} f(x) \ep^n
\label{eq:dd:est}
\eeq
for randomly selected directions $\ep^n\sim N(\bf{0},\sigma^2\Id)$ and the scalar directional derivatives along those directions.   We observe that the variance of the estimator is independent of $\sigma$, but keep this in the definition for consistency with the other approaches.   The DD estimator may also be viewed as an evolutionary process, with the each sample in \eqref{eq:dd:est} forming a member of the `swarm'.

\subsection{Analyzing the Directional Derivative Estimator}

If we draw vectors $\ep$ from a zero mean distribution with covariance $\sigma^2\Id$, we obtain the component
\begin{align}
\ave{\ep_iD_\ep f(x)}&= \ave{\ep_i\sum_j \ep_jg_j} =\sum_j \sigma^2\delta_{ij}g_j = \sigma^2 g_i
\end{align}
Hence, for $\ep$ sampled independently from a zero mean distribution with variance $\sigma^2$ 
\beq
\ave{\hat{g}_i} = \frac{1}{\sigma^2}\ave{\ep_iD_\ep f(x)} = g_i
\eeq
The DD gradient estimator is therefore unbiased. Its variance can be readily calculated as 
\beq
\ave{\hat{g}_i^2}- \ave{\hat{g}_i}^2 =\frac{1}{S}\br{g_i^2  + \sum_{j} g_j^2 }
\eeq

In GP we need to only compute $f$ once per sample, whereas in DD we need approximately two function evaluations (using Forward Mode AutoDiff).  For a fair comparison we assume that GP can use twice as many samples as the DD and AS sampler for their respective gradient estimators. Note, however, that even using twice as many samples, GP will typically result in a worse estimator of the gradient compared to DD or GP-AS.


\section{Efficient Communication\label{sec:eff:comm}}

A key insight in \cite{salimans17} is that the sampling process can be distributed across multiple machines, $i\in\{1,\ldots,S\}$ so that
\beq
f'(x) \approx \frac{1}{S\sigma^2}\sum_{i=1}^S {\ep^i f(x+\ep^i)}
\eeq
where $\ep^i$ is a vector sample and $i$ is the sample index. Each machine $i$ can separately calculate $f(x+\ep^i)$. Provided each machine $i$ knows the random seed used to generate the $\ep^j$ of each other machine, it therefore knows what all the $\ep^j$ are (by sampling according to the known seeds) and can thus calculate the SGD update $x^{new}$ based on communicating only the $S$ scalar values $f(x+\ep^i)$. That is, there is no requirement to send the vectors $\ep^i$ between machines (only the scalar values $f(x+\ep^i)$ need be sent), keeping the transmission costs very low. 

The same seed sharing approach can be used within the more general VO setting, as well as for the DD estimator. Thus both approaches are efficiently parallelizable.


\section{Demonstrations}\label{sec:demos}

\subsection{Quartic Objective}

To demonstrate the difference between the DD and GP approaches, we consider the function
\beq
f(x) = \frac{1}{D}\sum_{i=1}^D x_i^4
\eeq
This function has non-zero derivatives at third order, meaning that $\hat{g}_{GP}$ will be biased; note that in contrast $\hat{g}_{DD}$ is unbiased for any $f$.

\begin{figure}
\floatbox[{\capbeside}]{figure}[\FBwidth]
{\caption{Quartic $f$. $D=100$ dimensional $x$, with $S=5$ samples. $f(x)=\sum_i x_i^4/D$. The root mean squared error between $\hat{g}$ and the true gradient $g$. We plot the Gaussian Perturbation estimator in blue (with antithetic sampling in green) and the Directional Derivative estimator in red. For each value of $\sigma$ the $D$-dimensional inputs $x$ were sampled from a zero mean unit covariance Gaussian with the results presented as averages over 1000 experiments.  We also show the analytic approximations.}\label{fig:S5}}
{\scalebox{1.5}{\includegraphics[width=5cm]{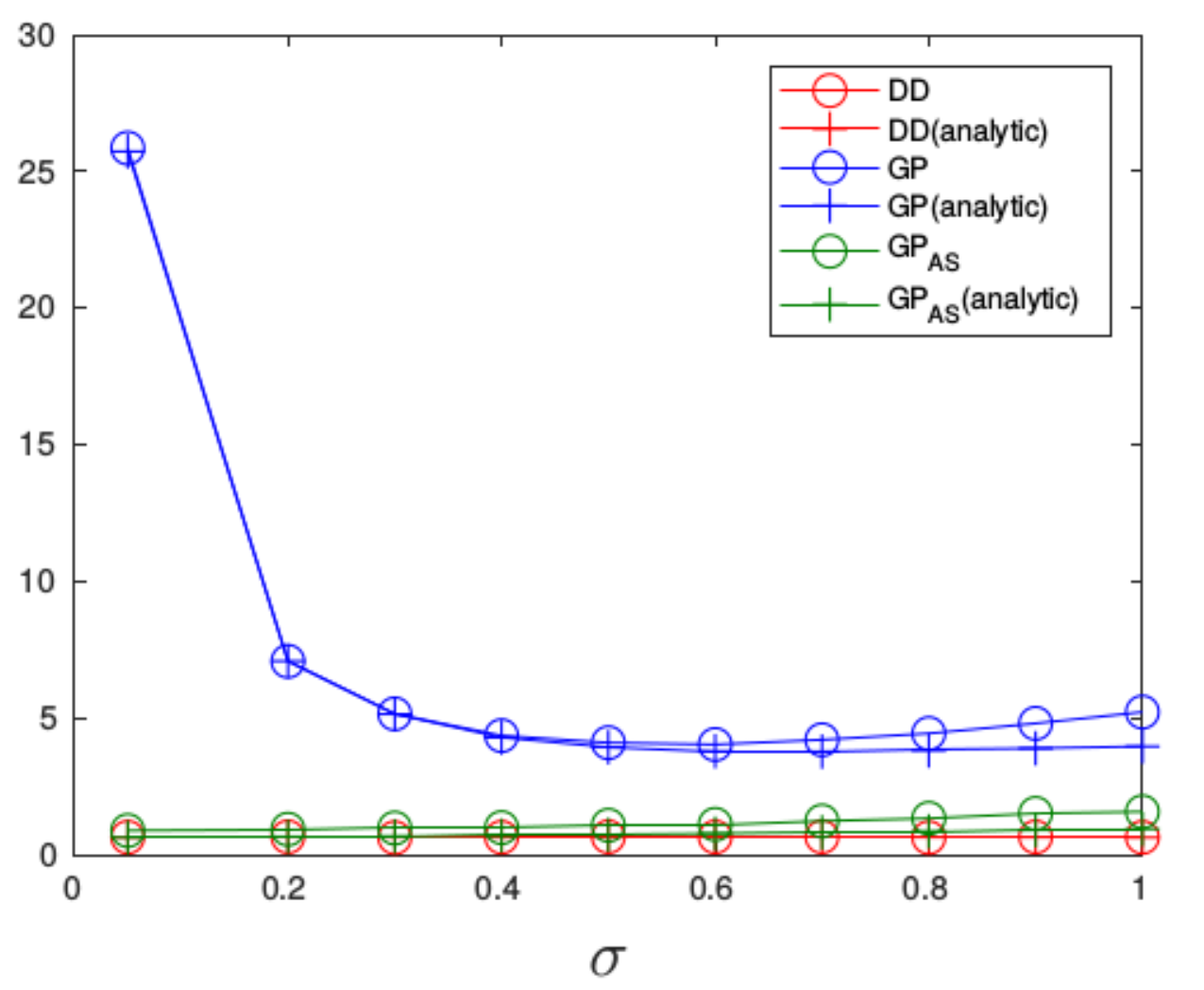}}}
\end{figure}

In \figref{fig:S5} we plot the (square root) of the squared error between the approximate gradient and true gradient (averaged over the $D$ dimensions). As we can see, the error for the DD approach is significantly lower than for the GP approach. As predicted by \eqref{eq:gp:variance}, the error grows dramatically as $\sigma^2$ reduces towards zero, and also grows roughly quadratically with increasing $\sigma$. 

This is in contrast to the GP-AS, which we see in \figref{fig:S5} lowers the error significantly. As we predicted (see \eqref{eq:gp:as:variance}) the antithetic sampling allows both the bias and variance of the estimator to tend to zero as we take $\sigma$ to zero. This results in the GP-AS error approaching that of the DD estimator as $\sigma$ decreases. \Figref{fig:S5} also shows that our analysis fits the experiments accurately in the regime of small $\sigma$ (where the Taylor expansion \eqref{eq:taylor} is accurate).  

\subsection{Neural Network Problem}

To understand how they behave in a setting of more practical relevance, we now examine the difference between the DD and GP approaches when applied to the optimization of a neural network. We also examine the use of antithetic sampling in the GP method.

We run experiments on the task of classifying MNIST digits, using each approach to approximating the gradient as input into the Adam  \cite{adam} optimization scheme (with fixed hyperparameters). We use a fully connected network with two hidden layers of size 300 and 100 respectively, and ReLu activation functions. This network is small relative to the sizes often used in deep learning, but it still contains hundreds of thousands of parameters and is highly non-linear. Therefore its loss surface will be largely representative of those seen in practice.

We perform the optimization across a range of values of $\sigma$, and simulate 1000 distributed workers performing the gradient approximations. This is equivalent to taking $S=1000$ in \eqref{eq:gp:est} and \eqref{eq:dd:est}. We use a baseline to reduce variance in the GP estimator, taking the difference of the loss from its moving average.

\begin{figure*}[ht]
		\subfigure[]{\includegraphics[width=0.45\tw]{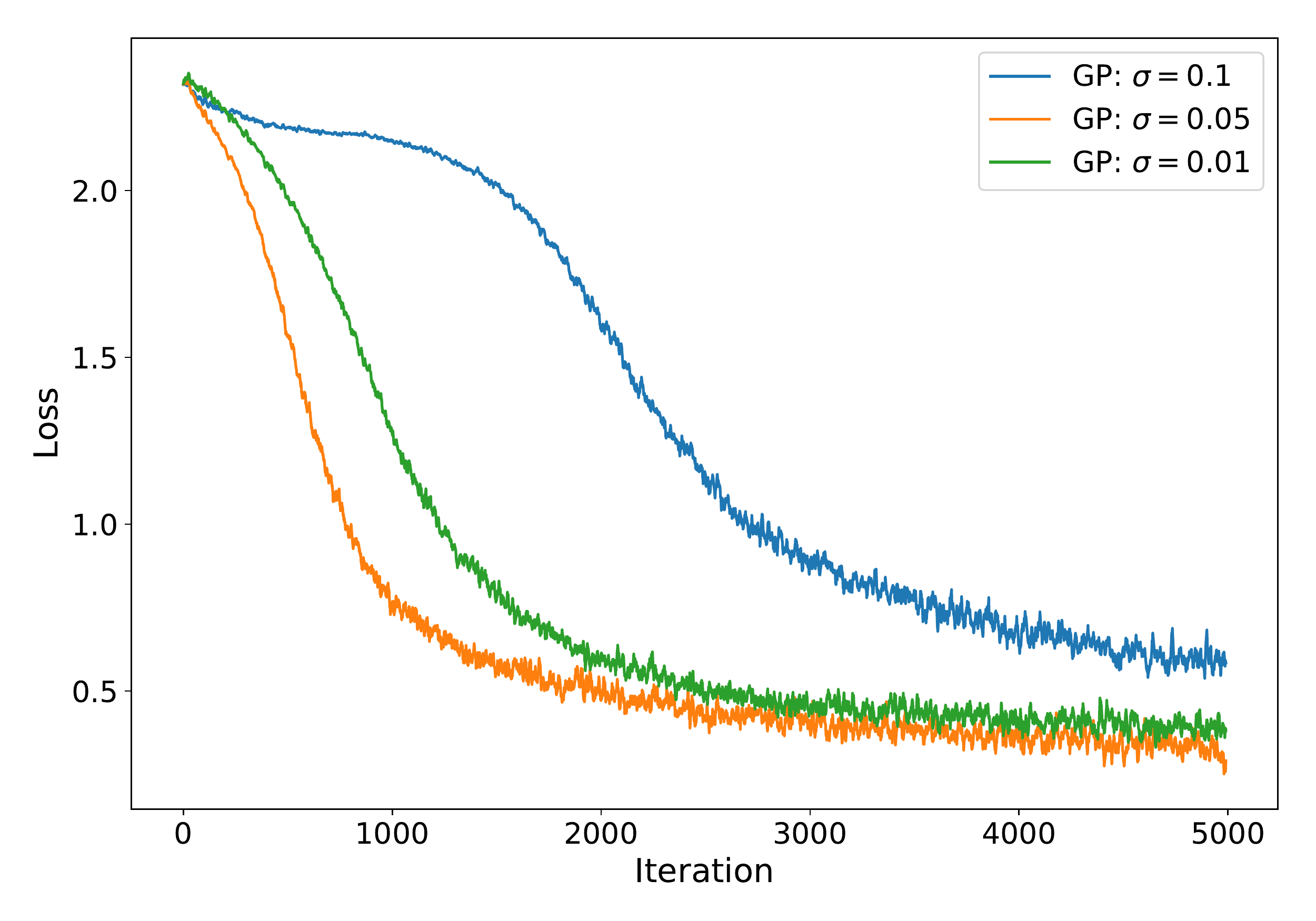}}\hcm
		\subfigure[]{\includegraphics[width=0.45\tw]{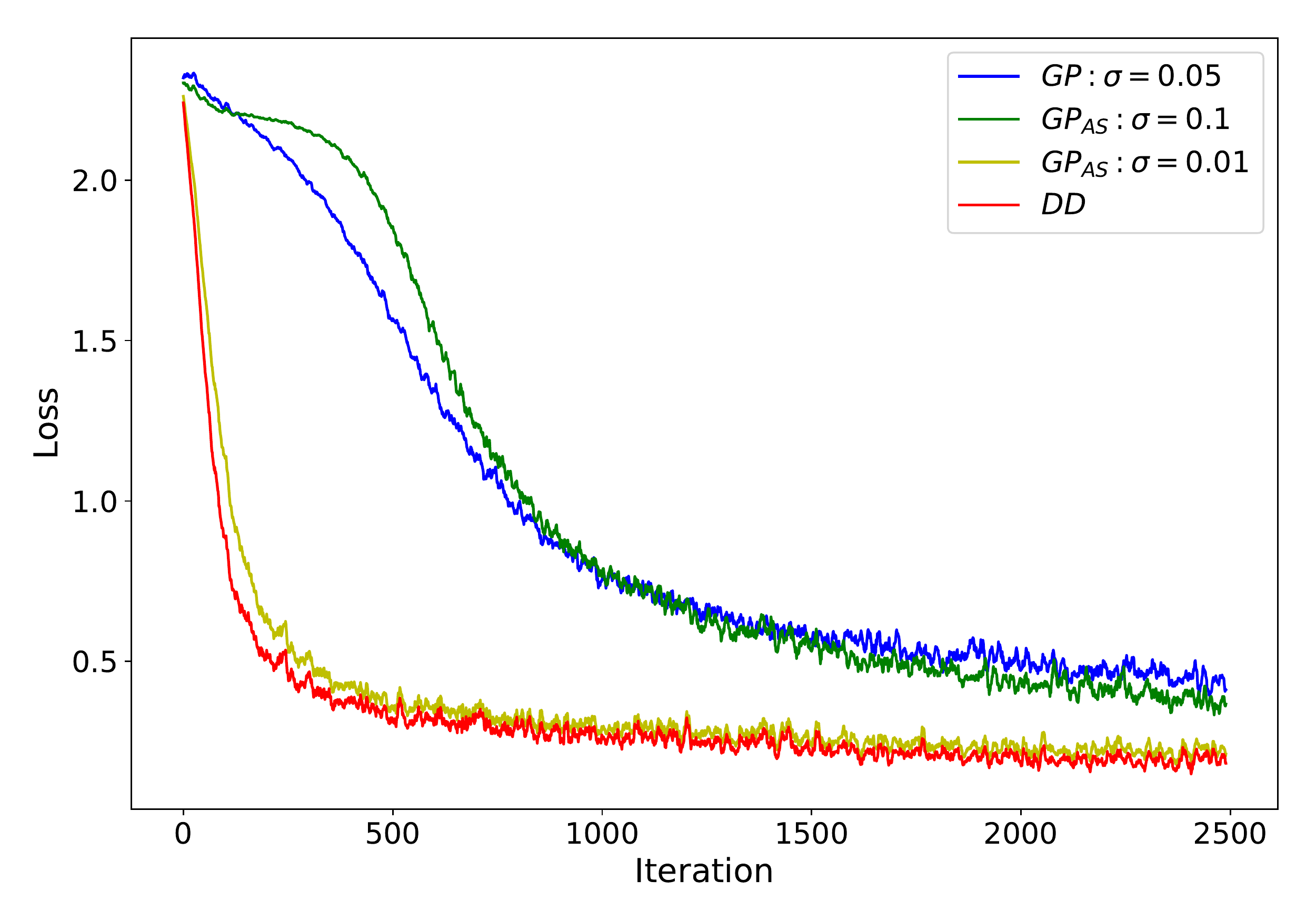}}
	\caption{The cross entropy loss of the neural network during training (taken as a moving average of the last 10 losses). (a) Using the Gaussian Perturbation estimator without antithetic sampling, for different values of $\sigma$. (b) Using the Gaussian Perturbation estimator with antithetic sampling (for two values of $\sigma$), as well as the Directional Derivative estimator. We also include a training sequence using Gaussian Perturbation without antithetic sampling for comparison  (at the near optimal value of $\sigma$). \label{fig:nn}}
\end{figure*}

In \figref{fig:nn}, we plot the loss of the network as it is trained using the various gradient estimators we have discussed. We observe from \figrefm{fig:nn}{a} that, as expected, using the GP approach without antithetic sampling results in relatively poor performance if $\sigma$ is too small or too large. This is attributable to the trade-off between bias and variance discussed earlier.

In \figrefm{fig:nn}{b}, we see that the DD method is far superior to using GP without antithetic sampling. We observe that antithetic sampling allows $\sigma$ to be decreased without impacting performance, validating our theoretical observations. This improves performance substantially in the GP method, and if $\sigma$ is sufficiently small then this method approaches the performance of the DD method.

\section{Discussion}


The high variance of the GP gradient estimator is also noted by \cite{wierstra14}. They observe that as the search distribution $p(x|\theta)$ narrows (during the optimization of both $\mu$ and $\sigma$ to try and minimize some loss surface), the variance of the updates increase. This is due to the bias/variance trade-off discussed previously. They use control variates and natural gradients to try to counter the high variance. 

Methods to reduce variance in gradient estimates are examined in \cite{greensmith}. They show that using a baseline can reduce the variance of Monte Carlo estimates for gradients (in Markov Decision Processes). However they note the optimal baseline may not be known. Our results have shown that a baseline can make the GP gradient estimator viable, as long as it is appropriately chosen.


\section{Conclusion\label{sec:conc}}

Stochastic Variational Optimization is an attractive approach for performing distributed optimization. In the case of using a Gaussian Perturbation, it is vital to use a variance reduction method to make the method practical, and either antithetic sampling or a baseline is particularly appropriate in this context. Alternatively one can use the SPSA estimator, which will have the same bias and similarly well-behaved variance as the GP estimator with antithetic sampling. These methods make it possible to perform distributed optimization efficiently.

For differentiable objectives, using Stochastic Directional Derivatives is preferable to Stochastic Variational Optimization. The variance of the estimator is generally superior and, compared to Variational Optimization, it is parameter free, meaning that there is no requirement to experimentally find a suitable variance $\sigma^2$ (as is the case for the variational approach). 

\clearpage
\newpage

\bibliographystyle{unsrt}
\bibliography{bib/optim}

\begin{thebibliography}{10}

\bibitem{zinkevich2010}
{Zinkevich, M. A.}, {Weimer, M.}, {Smola, A.}, and {Li, L.}
\newblock {Parallelized Stochastic Gradient Descent}.
\newblock In {\em Proceedings of the 23rd International Conference on Neural
  Information Processing Systems - Volume 2}, NIPS'10, pages 2595--2603, USA,
  2010.

\bibitem{dean2012}
{Dean, J.}, {Corrado, G. S.}, {Monga, R.}, {Chen, K.}, {Devin, M.}, {Le, Q.
  V.}, {Mao, M. Z.}, {Ranzato, M.}, {Senior, A.}, {Tucker, P.}, {Yang, K.}, and
  {Ng, A. Y.}
\newblock {Large Scale Distributed Deep Networks}.
\newblock In {\em Proceedings of the 25th International Conference on Neural
  Information Processing Systems - Volume 1}, NIPS'12, pages 1223--1231, USA,
  2012.

\bibitem{salimans17}
T.~{Salimans}, J.~{Ho}, X.~{Chen}, and I.~{Sutskever}.
\newblock {Evolution Strategies as a Scalable Alternative to Reinforcement
  Learning}.
\newblock {\em arXiv preprint arXiv:1703.03864}, March 2017.

\bibitem{spall}
J.~C. Spall.
\newblock Multivariate stochastic approximation using a simultaneous
  perturbation gradient approximation.
\newblock {\em IEEE Transactions On Automatic Control}, 37(3):332--341, 1992.

\bibitem{StainesBarber12}
J.~{Staines} and D.~{Barber}.
\newblock {Variational Optimization}.
\newblock {\em arXiv preprint arXiv:1212.4507}, December 2012.

\bibitem{Williams:1992:SSG:139611.139614}
R.~J. Williams.
\newblock {Simple Statistical Gradient-Following Algorithms for Connectionist
  Reinforcement Learning}.
\newblock {\em Mach. Learn.}, 8(3-4):229--256, May 1992.

\bibitem{StainesBarber13}
J.~Staines and D.~Barber.
\newblock {Optimization by Variational Bounding}.
\newblock In {\em 21st European Symposium on Artificial Neural Networks,
  {ESANN} 2013, Bruges, Belgium, April 24-26, 2013}, 2013.

\bibitem{NES}
D.~Wierstra, T.~Schaul, J.~Peters, and J.~Schmidhuber.
\newblock {Natural Evolution Strategies}.
\newblock In {\em CEC 2008}, pages 3381--3387, Piscataway, NJ, USA, June 2008.
  Max-Planck-Gesellschaft, IEEE.

\bibitem{hammersley}
{Hammersley, J. M.} and {Morton, K. W.}
\newblock {A New Monte Carlo Technique: Antithetic Variates}.
\newblock {\em Mathematical Proceedings of the Cambridge Philosophical
  Society}, 52(3):449–475, 1956.

\bibitem{fishman}
{Fishman, G. S.} and {Huang, B. D.}
\newblock {Antithetic Variates Revisited}.
\newblock {\em Commun. ACM}, 26(11):964--971, November 1983.

\bibitem{greensmith}
{Greensmith, E.}, {Bartlett, P. L.}, and {Baxter, J.}
\newblock {Variance Reduction Techniques for Gradient Estimates in
  Reinforcement Learning}.
\newblock {\em J. Mach. Learn. Res.}, 5:1471--1530, December 2004.

\bibitem{baydin15}
{Baydin A. G.}, {Pearlmutter, B. A.}, and {Radul, A. A.}
\newblock {Automatic Differentiation in Machine Learning: a Survey}.
\newblock {\em CoRR}, abs/1502.05767, 2015.

\bibitem{adam}
{Kingma D. P.} and {Ba J.}
\newblock {Adam: {A} Method for Stochastic Optimization}.
\newblock {\em CoRR}, abs/1412.6980, 2014.

\bibitem{wierstra14}
{Wierstra, D.}, {Schaul, T.}, {Glasmachers, T.}, {Sun, Y}, {Peters, J.}, and
  {Schmidhuber, J}.
\newblock {Natural Evolution Strategies}.
\newblock {\em J. Mach. Learn. Res.}, 15(1):949--980, January 2014.

\end{thebibliography}

\newpage
\onecolumn
\appendix

\section{Appendix: Bias and Variance Calculations\label{sec:details}}

\subsection{GP\label{sec:svo:details}}

\subsubsection{The bias}
Expanding \eqref{eq:gp:est} to leading orders in $\ep$ we have
\beq
\hat{g}_i \approx  \frac{1}{S\sigma^2}\sum_{n=1}^S \ep_i^n \br{f(x)+\sum_a \ep^n_a g_a + \frac{1}{2}\sum_{a,b} \ep^n_a \ep^n_b H_{ab} + \frac{1}{3!}\sum_{abc} \ep^n_a \ep^n_b \ep^n_c I_{abc}}
\eeq
  
Since the Gaussian is symmetric, taking expectations with respect to Gaussian $u$ gives
\beq
\ave{\hat{g}_i} \approx g_i+\sett{I}_i\sigma^2
\eeq
where
\beq
\sett{I}_i\equiv \frac{1}{3!}\br{3I_{iii}+\sum_{a\neq i}\br{I_{iaa}+I_{aia}+I_{aai}}}
\eeq
Due to the symmetry of partial derivatives, this can be written
\beq
\sett{I}_i = \frac{1}{2}\br{I_{iii}+\sum_{a\neq i}I_{iaa}}
\eeq
Except in special cases (for example $f$ has zero third and higher order derivatives), $\hat{g}_i$ is therefore not an unbiased estimator of the gradient $g_i$. 

\subsubsection{The variance}
To approximate the variance of the Gaussian perturbation estimator, we consider
\beq
\hat{g}_i \approx  \frac{1}{S\sigma^2} \sum_n \ep^n_i \br{f(x)+\sum_j \ep^n_j g_j+\half\sum_{kl}\ep^n_k\ep^n_lH_{kl} + \frac{1}{3!}\sum_{abc}\ep^n_a\ep^n_b\ep^n_cI_{abc}}
\label{eq:gp:var:calc}
\eeq
Using this we can approximate $\ave{\hat{g}_i^2}$ as

\begin{align}
    \frac{1}{S^2\sigma^4} \sum_{m,n} \mathbb{E}\bigg[ \ep^m_i\ep^n_i &\bigg( f(x) + \sum_j\ep^n_jg_j + \half\sum_{kl}\ep^n_k\ep^n_lH_{kl}+ \frac{1}{3!}\sum_{abc} \ep^n_a \ep^n_b \ep^n_c I_{abc} \bigg) \nonumber \\
     &\bigg( f(x) + \sum_j\ep^m_jg_j + \half\sum_{kl}\ep^m_k\ep^m_lH_{kl}+ \frac{1}{3!}\sum_{abc} \ep^m_a \ep^m_b \ep^m_c I_{abc} \bigg) \bigg] \nonumber
\end{align}
\begin{align}
    =\frac{1}{S^2\sigma^4} &\bigg( f^2\sigma^2S+S\sigma^4\sum_{j \neq i} g_j^2+3S\sigma^4g_i^2+(S^2-S)\sigma^4g_i^2 \nonumber \\
    &+ fS\br{3\sigma^4H_{ii}+\sigma^4\trace{H}-\sigma^4H_{ii}}
    + \frac{\tilde{H}_i}{4} + \frac{\tilde{I}_i}{3} + O(\ep^8) \bigg)
\end{align}

\text{The term $\tilde{H}_i$ is given by}
\begin{align}
\tilde{H}_i\equiv &\sum_{mnklab}H_{kl}H_{ab}\ave{\ep^m_i\ep^n_i\ep^n_k\ep^n_l\ep^m_a\ep^m_b} \\
&= S\sum_{klab}H_{kl}H_{ab}\ave{\ep_i\ep_i\ep_k\ep_l\ep_a\ep_b}+ (S^2-S)\sum_{klab}H_{kl}H_{ab}\ave{\ep_i\ep_a\ep_b}\ave{\epp_i\epp_k\epp_l}
\end{align}
Due to symmetry of the Gaussian, the final term (where $\ep$ and $\epp$ are independent Gaussian random variables) in the above expression is zero. If we consider for simplicity that the Hessian is diagonal, then
\begin{align}
\sum_{klab}&H_{kl}H_{ab}\ave{\ep_i^2\ep_k\ep_l\ep_a\ep_b}= \nonumber \\ 
&\ave{\ep^6}H_{ii}^2 +\ave{\ep^4}\ave{\ep^2}\br{2H_{ii}\sum_{a\neq i}H_{aa} + \sum_{a\neq i}H_{aa}^2}+\ave{\ep^2}^3\br{\sum_{a\neq i}H_{aa}}^2
\end{align}
\text{which for the Gaussian gives}
\begin{align}
\tilde{H}_i &= S\sum_{klab}H_{kl}H_{ab}\ave{\ep_i^2\ep_k\ep_l\ep_a\ep_b}\equiv S\sigma^6\sett{H}_i \\
\sett{H}_i&\equiv 15H_{ii}^2+6H_{ii}\sum_{a\neq i}H_{aa}+3\sum_{a\neq i}H_{aa}^2+\sum_{a,b \neq i, a \neq b}H_{aa}H_{bb}
\end{align}

The term $\tilde{I}_i$ follows a similar calculation
\begin{align}
    \tilde{I}_i = f(x)\sum_{mnabc}\ave{\ep^m_i\ep^n_i\ep^m_a \ep^m_b \ep^m_c} I_{abc} + \sum_{mnabcj}\ave{\ep^m_i\ep^n_i\ep^n_j\ep^m_a \ep^m_b \ep^m_c} g_j I_{abc}
\end{align}

The first term goes to zero by symmetry of the Gaussian, and thus
\begin{align}
    \tilde{I}_i &= S\sum_{abcj}\ave{\ep_i^2\ep_j\ep_a \ep_b \ep_c} g_j I_{abc} + (S^2-S)\sum_{abcj}\ave{\ep_i\ep_a \ep_b \ep_c} \ave{\tilde{\ep_i}\tilde{\ep_j}} g_j I_{abc} \\
    &= 3S\sigma^6 \bigg( 5g_i I_{iii} +
    \begin{aligned}[t]
    &3\sum_{a \neq i}\big[g_iI_{iaa} + g_aI_{iia} \big] + \sum_{a,b \neq i}g_aI_{abb}  \bigg) \nonumber \\ 
    &+ 3\sigma^6 g_i(S^2-S)\bigg(I_{iii} + \sum_{a \neq i}I_{iaa}\bigg)
    \end{aligned}\\
    &= 3S\sigma^6 (\sett{J}_i + 2Sg_i\sett{I}_i
)    
\end{align}
Where $\sett{I}_i$ is as above, and we define
\begin{equation}
\sett{J}_i = 4g_i I_{iii} + \sum_{a \neq i}\big[2g_iI_{iaa} + 3g_aI_{iia} \big] + \sum_{a,b \neq i}g_aI_{abb}
\end{equation}

Thus the variance of this estimator is
\beq
\ave{\hat{g}_i^2}- \ave{\hat{g}_i}^2=
\frac{1}{S} \br{\frac{f^2}{\sigma^2}+\sum_{j} g_j^2+g_i^2+f\br{\trace{H}+2H_{ii}}+ \sigma^2 \bigg( \frac{\sett{H}_i}{4} + \sett{J}_i\bigg)}+O\br{\sigma^4}
\label{eq:gp:variance:details}
\eeq
For a large dimension $D\gg 1$ we can approximate this by
\beq
\ave{\hat{g}_i^2}- \ave{\hat{g}_i}^2=
\frac{1}{S} \br{\frac{f^2}{\sigma^2}+D\sett{G}_2 +Df\sett{H}+ \sigma^2 \bigg( \frac{\sett{H}_i}{4} + \sett{J}_i\bigg)}+O\br{\sigma^4}
\label{eq:gp:variance:app}
\eeq
where $\sett{G}_2$ is the average squared gradient $\sum_{j=1}^D g_j^2/D$ and $\sett{H}$ is the average second derivative  $\sum_{j=1}^D H_{jj}/D$. 

The squared error is given by 
\beq
\ave{ \br{\hat{g}_i - g_i}^2} = \ave{\br{\hat{g}_i}^2} - 2g_i\ave{\hat{g}_i} + g_i^2
\eeq
Up to $O(\sigma^2)$ this has the same value as the variance \eqref{eq:gp:variance:details}.

\subsubsection{The Quadratic Objective\label{sec:gp:quad}}

All smooth functions will look quadratic around a local minimum. An important canonical objective is therefore the quadratic
\beq
f(x) = \frac{1}{2D}\sum_{i=1}^D x_i^2
\label{eq:quadf}
\eeq
In this case $g_i = x_i/D$ and $H_{ij}=\delta_{ij}/D$. All third and higher order derivatives are zero. For this setting we can exactly calculate the GP error, giving the expression
	\beq
	\ave{\hat{g}_i^2}- \ave{\hat{g}_i}^2=
	\frac{1}{S} \br{\frac{f^2}{\sigma^2}+\sum_{j} g_j^2+g_i^2+f\br{\trace{H}+2H_{ii}}+ \frac{\sigma^2}{4}\sett{H}_i}
	\label{eq:gp:variance:quad}
	\eeq
where
\beq
\sett{H}_i\equiv 15H_{ii}^2+6H_{ii}\sum_{a\neq i}H_{aa}+3\sum_{a\neq i}H_{aa}^2+\br{\sum_{a\neq i}H_{aa}}^2
\eeq
which gives
\beq
D^2\sett{H}_i = 15+6(D-1)+3(D-1)+(D-1)^2 = 7+7D+D^2
\eeq
For $D\gg 1$ we have therefore
\beq
\ave{\hat{g}_i^2}- \ave{\hat{g}_i}^2=
\frac{1}{S} \br{\frac{f^2}{\sigma^2}+f+ \frac{\sigma^2}{4}} + O\br{\frac{1}{DS}}
\label{eq:gp:variance:quad:simple}
\eeq

\subsection{Directional Derivative\label{sec:dd:details}}
Based on $S$ samples, we can form an estimator for the gradient using
\beq
\hat{g}_i =  \frac{1}{S\sigma^2}\sum_{n=1}^S \ep^n_i D_{\ep^n}f(x) = \frac{1}{S\sigma^2}\sum_{n=1}^S\sum_j \ep_i^n \ep_j^n g_j
\eeq
This gradient estimator is unbiased, $\ave{\hat{g}_i}=g_i$ and its variance can be readily calculated as
\beq
\ave{\hat{g}_i^2} = \frac{1}{S^2\sigma^4}\sum_{m,n}\sum_{j,k} g_j g_k\ave{\ep_i^n \ep_j^n\ep_i^m \ep_k^m}
\eeq
We can use the result that $\ave{\ep_i^n \ep_j^n\ep_i^m \ep_k^m}$ is
\beq
3\sigma^4\delta_{mn}\delta_{ij}\delta_{ik}
+ \sigma^4\delta_{mn}\delta_{jk}\br{1-\delta_{ij}}+\sigma^4(1-\delta_{mn})\delta_{ij}\delta_{ik}
\eeq
to give
\beq
\ave{\hat{g}_i^2}- \ave{\hat{g}_i}^2 =\frac{1}{S}\br{g_i^2  + \sum_{j} g_j^2 }
\eeq

\end{document}